\documentclass[runningheads]{llncs}
\usepackage[T1]{fontenc}
\usepackage{graphicx}
\usepackage{amsmath}
\begin{document}
\title{Beyond Accuracy: An Explainability-Driven Analysis of Harmful Content Detection}
\titlerunning{Explainability-Driven Analysis of Harmful Content Detection}
%
\author{Trishita Dhara\inst{1}\orcidID{0000-0002-9325-9286} \and
Siddhesh Sheth\inst{2}\orcidID{0009-0003-7154-0477} }
\authorrunning{T. Dhara and S. Sheth}
%
\institute{
Upper Hand, Indianapolis, IN 46204, USA\\
\email{trishitadhara123@gmail.com}
\and
Ace Rent a Car, Indianapolis, IN 46204, USA\\
\email{shethsiddhesh268@gmail.com}
}

\maketitle              
\begin{abstract}
Although automated harmful content detection systems are frequently used to monitor online platforms, moderators and end users frequently cannot understand the logic underlying their predictions. While recent studies have focused on increasing classification accuracy, little focus has been placed on comprehending why neural models identify content as harmful, especially when it comes to borderline, contextual, and politically sensitive situations. In this work, a neural harmful content detection model trained on the Civil Comments dataset is analyzed explainability-drivenly. Two popular post-hoc explanation methods, Shapley Additive Explanations and Integrated Gradients, are used to analyze the behavior of a RoBERTa-based classifier in both correct predictions and systematic failure cases. Despite strong overall performance, with an area under the curve of 0.93 and an accuracy of 0.94, the analysis reveals limitations that are not observable from aggregate evaluation metrics alone. Integrated Gradients appear to extract more diffuse contextual attributions while Shapley Additive Explanations extract more focused attributions on explicit lexical cues. The consequent divergence in their outputs manifests in both false negatives and false positives. Qualitative case studies reveal recurring failure modes such as indirect toxicity, lexical over-attribution, or political discourse. The results suggest that explainable AI can foster human-in-the-loop moderation by exposing model uncertainty and increasing the interpretable rationale behind automated decisions. Most importantly, this work highlights the role of explainability as a transparency and diagnostic resource for online harmful content detection systems rather than as a performance-enhancing lever.

\end{abstract}
\section{Introduction}

In order to detect and control harmful content, such as hate speech, harassment, and toxic language, online platforms are depending more and more on automated systems. Machine learning models have become essential to content moderation workflows as the amount of user-generated content keeps increasing because they allow for the quick and extensive identification of potentially hazardous content. The detection performance across established benchmarks has significantly improved due to recent developments in pretrained language models ~\cite{ref_roberta,ref_bert}.

Despite recent advances, harmful content detection remains a challenging problem with significant ethical and social implications. False negatives allow abusive or harmful behavior to persist, whereas false positives risk suppressing legitimate expression. As a result, accuracy based evaluation in itself fails to provide a complete view of system performance in real-world moderation settings.  Human moderators must have access to interpretable rationales underlying model decisions in order to assess accountability, consistency, and fairness of the system. In the absence of such transparency, the responsible deployment of content moderation systems becomes difficult.

Explainable artificial intelligence (XAI) helps to show why and how a system made a certain decision. It has been a popular research area in recent years, to make machine learning models more trustworthy and transparent. Post-hoc XAI methods like Shapley Additive Explanations~\cite{ref_shap} and Integrated Gradients~\cite{ref_ig}, explain a model’s predictions by showing how each input feature influenced the result without modifying the underlying model architecture. Although these techniques are increasingly adopted into decision-support tools, structured analysis of their behavior and reliability remains limited, particularly in the domain of harmful content moderation.
Previous research in the domain of harmful content detection mostly focus on  improving classification recall through new models, better pretraining strategies, or extending datasets~\cite{ref_civilcomments,ref_toxicmodels}. Even though explanation methods have been applied to text classification in general, relatively less work studies how the explanation patterns varies between correct and incorrect predictions in harmful content detection. The study addresses this gap by implementing an explanation-based analysis of a harmful content detection model. A RoBERTa-based classifier~\cite{ref_roberta} is  fine-tuned on the Civil Comments dataset~\cite{ref_civilcomments} and then examined using  Shapley Additive Explanations and Integrated Gradients to understand model predictions. The analysis focuses on interpreting the model outputs rather than proposing new architecture, particular attention is given to ambiguous and context-sensitive cases commonly encountered in moderation practice. Both quantitative and qualitative assessments were conducted to understand failure modes of transformer based harmful content detection models. The rest of the paper is organized into the following sections - related works, methodology, results, discussion and conclusion, and limitations. 

\section{Related Work}

\subsection{Harmful Content and Toxic Language Detection}

A lot of research has been done on automated detection of toxic and harmful language as online platforms try to filter vast amounts of user-generated content. Early approaches relied on traditional machine learning models with manually constructed features, like lexical and syntactic cues~\cite{ref_icwsm,ref_naacl}. When deep learning was introduced, neural architectures such as convolutional and recurrent networks performed better on toxicity and hate speech benchmarks~\cite{ref_toxicmodels,ref_epj}. 

Pretrained transformer-based language models have become the most popular method for detecting toxic content in more recent research. By accurately simulating contextual semantics, models such as BERT and RoBERTa produce impressive results on a variety of datasets~\cite{ref_roberta,ref_bert}. Large-scale benchmarks, like the Civil Comments dataset, have made it possible to systematically analyze toxicity classifiers' bias, robustness, and generalization~\cite{ref_civilcomments,ref_bias}. However, it is still very difficult to correctly identify subtle and context-dependent forms of toxicity, such as political discourse and implicit harassment.

\subsection{Explainable Artificial Intelligence for Text Classification}

In order to improve machine learning systems' transparency and credibility, explainable artificial intelligence has emerged as a key area of study. Interpreting model predictions without changing the underlying architectures is the goal of post-hoc explanation techniques. Earlier works in explainability have shown that these explanations can assist in exposing false correlations and latent artifacts of the data, but are also vulnerable to model sensitivity due to instabilities, input perturbations and uncertainty changes indicated by perplexity shifts~\cite{ref_attention}. The faithfulness and reliability of post-hoc methods used for explanations on deep neural models still remain debated and investigated.  

\subsection{Explainability in Harmful Content Moderation}

Explainability has been studied extensively with natural language processing research works in general but not specifically applied to harmful content moderation. Most of the works examine explanation methods as audit mechanisms that detect bias in toxicity classifiers that focus mainly on identity terms and demographic attributes ~\cite{ref_civilcomments,ref_bias}. Some other works have introduced explanation-oriented moderation interfaces that help human moderators understand automated decisions in content classification tasks~\cite{ref_interfaces}. 

Despite these works, most of literature has approached question of explanation methods independently or has mainly emphasized fairness-driven goals, instead of attempts to systematically analyze and characterize explanation behavior on both successes and diagnostic failure cases. There seem to be little empirical studies from the comparative view of explanation methods that are related to false positives and false negatives in the harmful content identification task. This work builds on preceding studies by presenting a comparative study of Shapley Additive Explanations and Integrated Gradients with a neural toxicity classifier that helps extract insights regarding the behavior of these explanation methods and their errors in content moderation task.  

\section{Methodolody}

\subsection{Task Description}

A binary toxic content identification task is formed where each input text is either identified as a toxic or a non-toxic class. The binary labels are obtained from the continuous toxicity scores of Civil Comments dataset~\cite{ref_civilcomments} . This binary labeling is achieved from a fixed threshold of 0.5 such that text inputs identified with a toxicity score greater than or equal to 0.5 are considered as toxic. This thresholding is similar to numerous models in previous studies. This task definition fits the normal paradigm of real-world scenarios where automated moderation systems are required to perform binary classification on users' contents
\subsection{Dataset}

Civil Comments dataset~\cite{ref_civilcomments} is public comments with its metadata from 50 news websites across the globe. This dataset has been used in previous works to assess the robustness of models detecting harmful content and to study bias in automatic moderation systems~\cite{ref_civilcomments,ref_bias}. In the experiments, a subset of the dataset is used with 20,000 training instances, 4,000 validation instances and 4,000 test instances to be able to train a baseline model but keeping the costs reasonably low and the explanation behavior examinable. Although the reported explainability patterns are derived from a subset of the dataset, they capture model behavior shaped by learned lexical and contextual representations. Hence, these patterns are expected to generalize to larger samples and to other similar toxicity benchmarks. Extending the analysis to full-scale datasets remains a potential direction for future work.

\subsection{Model Architecture}

RoBERTa-base~\cite{ref_roberta} is selected as the classification model for this study. The model is a transformer-based language model that builds upon the BERT architecture~\cite{ref_bert} through optimized pretraining strategies, including dynamic masking and training on substantially larger text corpora. Its strong empirical performance across a wide range of natural language processing tasks makes it a suitable baseline for explainability analysis without introducing additional architectural complexity. A linear classification head is attached to the final hidden representation corresponding to the special classification token, and the full model is fine-tuned end-to-end for binary toxic content classification.

\subsection{Pipeline Overview}

\begin{figure}
\centering
\includegraphics[width=8 cm]{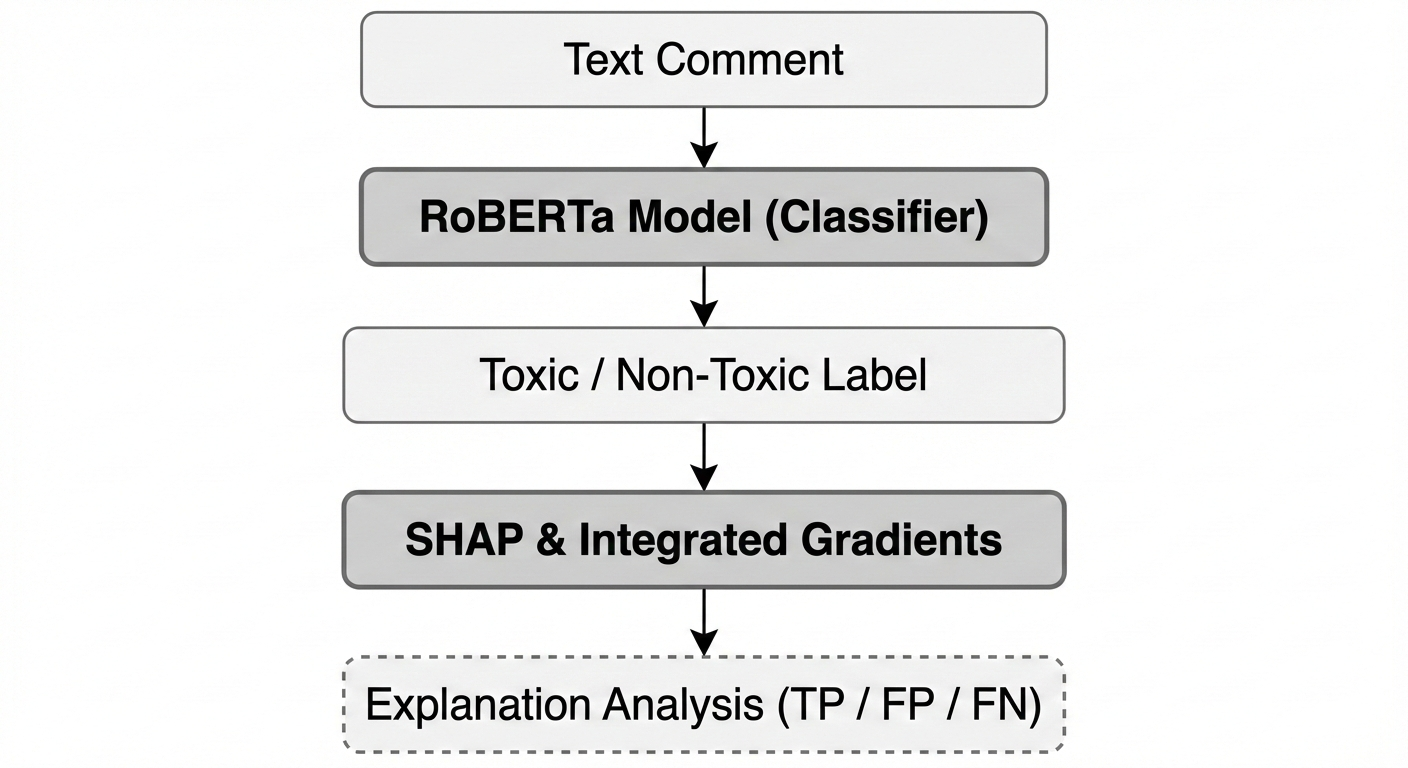}
\caption{Overview of the harmful content detection and explainability analysis pipeline.}
\label{fig:pipeline}
\end{figure}

Figure~\ref{fig:pipeline} illustrates the analysis pipeline employed in this work. Raw text inputs are first processed by the RoBERTa-based classifier to produce binary toxicity predictions. Next, post-hoc explanation techniques, namely Shapley Additive Explanations and Integrated Gradients, are executed on the model after training to yield token-level attribution score for individual model predictions. The model explanations for true positives, false positives, and false negatives are analyzed to discover any patterns, and failure modes that are consistent and applicable to automated content moderation.

This pipeline implements a practical moderation workflow where automated predictions are enhanced with explanations to facilitate human moderation and auditing.

\subsection{Training Configuration}

Model fine-tuning is performed using the AdamW optimizer with a batch size of 32 and a learning rate of $2 \times 10^{-5}$. The model is trained for one epoch, which is enough to achieve good prediction quality and prevents overfitting. The performance of the model is evaluated at the end of the training epoch in the validation set.

Harmful content evaluation metrics are defined on accuracy, precision, recall, F1-score, and area under the receiver operating characteristic curve for the toxic class on a held-out test set. These metrics are used in harmful content detection settings as they reflect the trade-offs between false positive and false negative errors while addressing class imbalance~\cite{ref_civilcomments,ref_toxicmodels}.

\subsection{Explainability Methods}

To understand model behavior beyond just overall performance measures, two post-hoc explainability techniques are applied: Shapley Additive Explanations and Integrated Gradients.

\paragraph{Shapley Additive Explanations.}
Shapley Additive Explanations are inspired by cooperative game theory and estimate the contribution of each input token to the prediction of a model~\cite{ref_shap}. As a model-agnostic approach, the method yields token-level attributions that facilitate direct interpretation of individual predictions.

\paragraph{Integrated Gradients.}
Integrated Gradients calculates token attributions by integrating gradients along a continuous path from a baseline input to the final input~\cite{ref_ig}. Unlike the Shapley Additive Explanations, this is a model-specific measure that retrieves the attribution signals spread throughout the input sequence, and therefore is affected by the wider contextual interactions.

\subsection{Explainability Evaluation Protocol}

In line with previous interpretability research~\cite{ref_lime,ref_interpret1}, a qualitative, case-based evaluation approach is favored over fully automated faithfulness metrics. Representative cases from the different prediction classes, true positives, false positives and false negatives, are compared to the explanation behavior of the model. This allows the detection of salient explanation patterns and failure modes that can be directly connected to real-life moderation scenarios.

\section{Results}

\subsection{Classification Performance}

The performance of the RoBERTa-based classifier is first evaluated on the previously mentioned Civil Comments test set. Table~\ref{tab:performance} summarizes the classification results using accuracy, precision, recall, and F1-score for the toxic class, along with the area under the receiver operating characteristic curve.

\begin{table}
\caption{Classification performance on the Civil Comments test set.}
\label{tab:performance}
\centering
\begin{tabular}{|l|c|}
\hline
Metric & Value \\
\hline
Accuracy & 0.9405 \\
AUC & 0.9371 \\
Precision (Toxic) & 0.6242 \\
Recall (Toxic) & 0.6203 \\
F1-score (Toxic) & 0.6222 \\
\hline
\end{tabular}
\end{table}

Although training is limited to a single epoch, the model attains strong overall performance, achieving an accuracy of 0.94 and an AUC of 0.94. These results are consistent with prior work demonstrating that pretrained transformer models provide effective baselines for harmful content detection~\cite{ref_roberta,ref_bert}. Performance on the toxic class is comparatively lower, with an F1-score of 0.62, highlighting the challenges posed by class imbalance and by context-dependent or implicitly expressed toxicity~\cite{ref_civilcomments,ref_toxicmodels}.

The objective of this study is not to optimize classification performance, but to establish a reliable baseline for explainability-focused analysis. The observed performance ensures that subsequent interpretation results are grounded in the behavior of a competitive moderation model, rather than artifacts of an underperforming system.

\subsection{Confusion Matrix Analysis}

\begin{figure}
\centering
\includegraphics[width=0.7\linewidth]{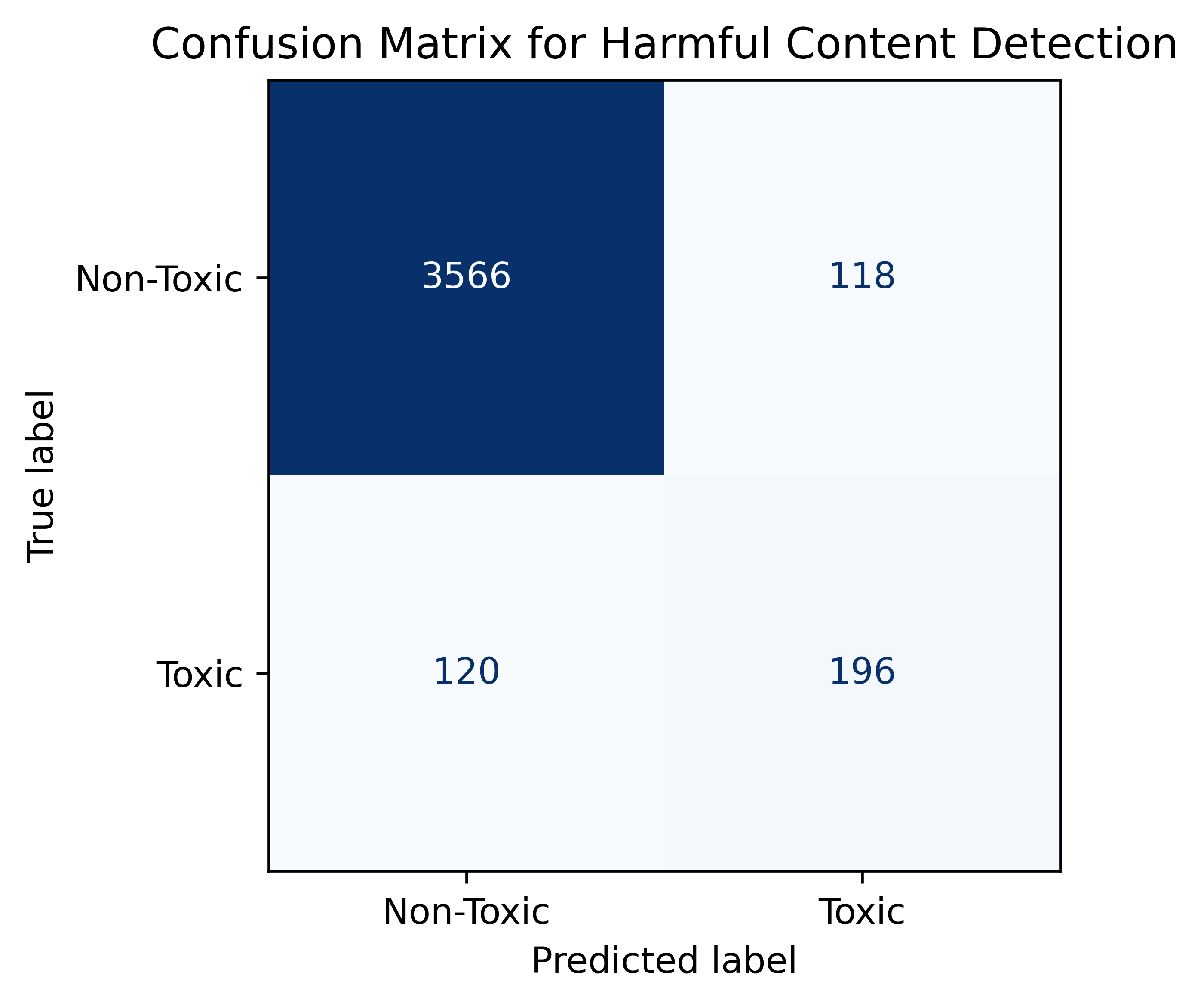}
\caption{Confusion matrix for binary harmful content classification on the Civil Comments test set.}
\label{fig:confusion}
\end{figure}

Figure~\ref{fig:confusion} presents the confusion matrix for binary toxicity classification on the test set. The model correctly predicts 3{,}566 examples out of the 4{,}000 test samples. The total false positives (118) and false negatives (120) are almost same in number, which shows that the model does not bias toward either over-flagging or under-detecting toxic content.

A neutral error pattern forms an optimal rationale for structured and comparative explanation behavior assessment across prediction groups. The false positive and false negative types, in particular, are vastly different moderation threads – undue moderation versus unfiltered moderation of harmful content. This makes them particularly significant for the human explainability assessment in human-in-the-loop~\cite{ref_interfaces} scenarios.

\subsection{Error Category Breakdown}

To facilitate a fine-grained analysis of the behavior of the model, its predictions on the test set are further classified into four classes: true positives, true negatives, false positives, and false negatives. The distribution is illustrated in Figure~\ref{fig:confusion}. False positives are comments with no harassment but politically affiliated or emotionally expressive, which strongly lexicalize the relevant signals. In turn, false negatives represent comments with indirect, rhetorical, or context-dependent toxicity, which indicate the harmfulness of the content without being explicit about it. Such patterns are in line with the previous research demonstrating that the surface lexicon is not always helpful for the detection of more subtle forms of toxicity~\cite{ref_civilcomments,ref_bias}.

\subsection{Implications for Explainability Analysis}

The quantitative findings call for a deeper analysis of the model explanations. Although performance scores achieved in the aggregate provide evidence of effective generalization and high predictive power, they do not help to understand why predictions have failed in particular scenarios. To this end, post-hoc explainability methods, as described in the analysis pipeline of Figure~\ref{fig:pipeline} are resorted to representative examples for each error type to reveal any possible systematic attribution trends and recurring failure scenarios which are missed by standard analysis procedures. 

\section{Explainability Analysis}

To characterize model behavior beyond summary metrics, post-hoc explainability methods are applied at the level of individual predictions. Shapley Additive Explanations and Integrated Gradients are used to generate token-level attribution scores for representative examples drawn from true positive, false positive, and false negative cases, following the pipeline shown in Figure~\ref{fig:pipeline}. A confidence-aware selection strategy is employed to focus the analysis on systematic model behavior rather than borderline or ambiguous predictions.

\subsection{Shapley Additive Explanations}

\begin{figure}
\centering
\includegraphics[width=9 cm]{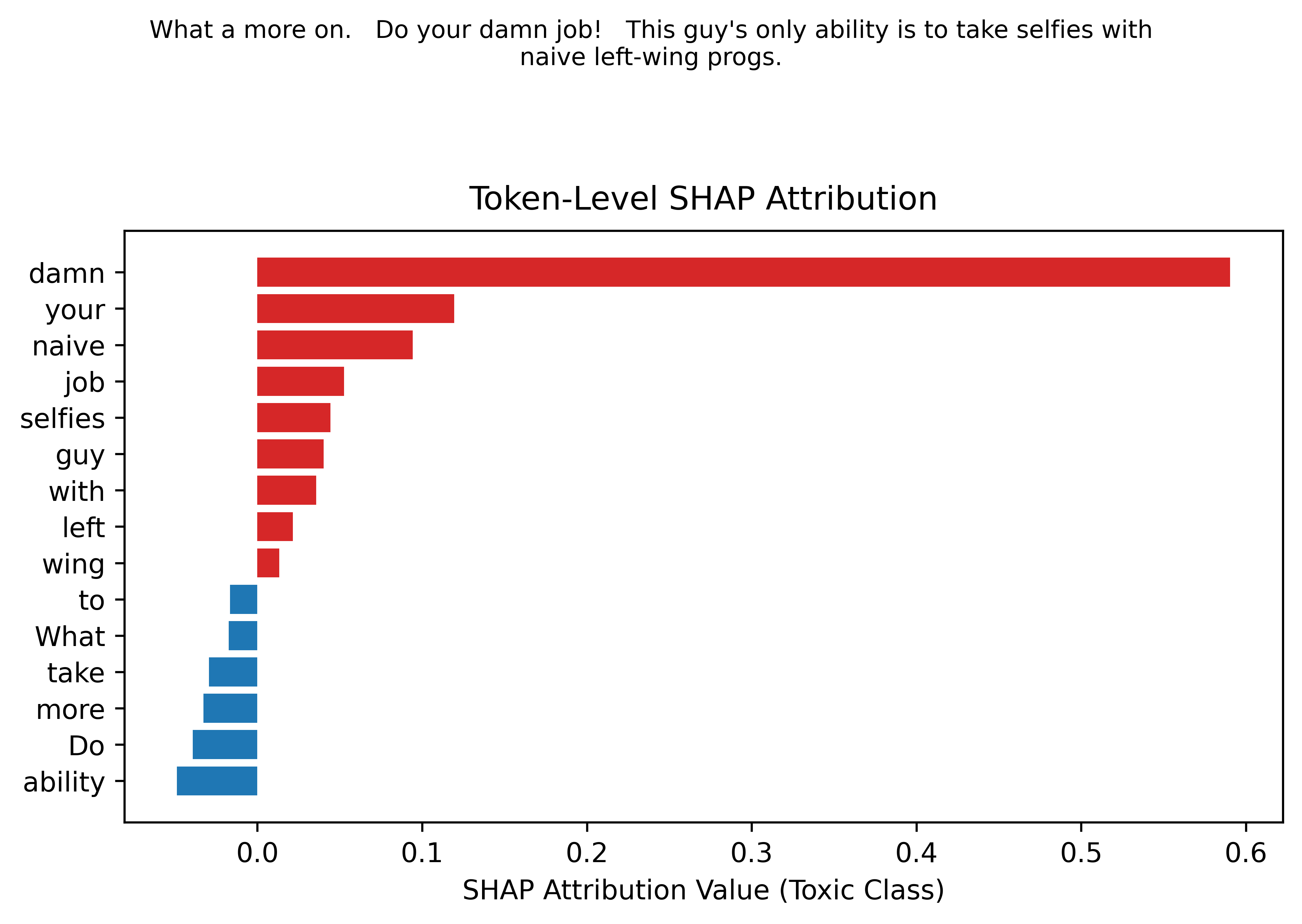}
\caption{Shapley Additive Explanations for a correctly classified toxic comment.}
\label{fig:shap_tp}
\end{figure}

Shapley Additive Explanations provide sparse, token-level attributions that highlight the most influential lexical cues contributing to the model’s prediction. For positive, correctly classified toxic examples, Shapley Additive Explanations always provided strong positive attribution scores to explicit insult terms and emotionally-laden vocabulary. An example can be seen in Figure~\ref{fig:shap_tp} where a minority of tokens drove the model’s prediction, leading to intuitive and interpretable explanations.

\begin{figure}[!ht]
\centering
\includegraphics[width=9 cm]{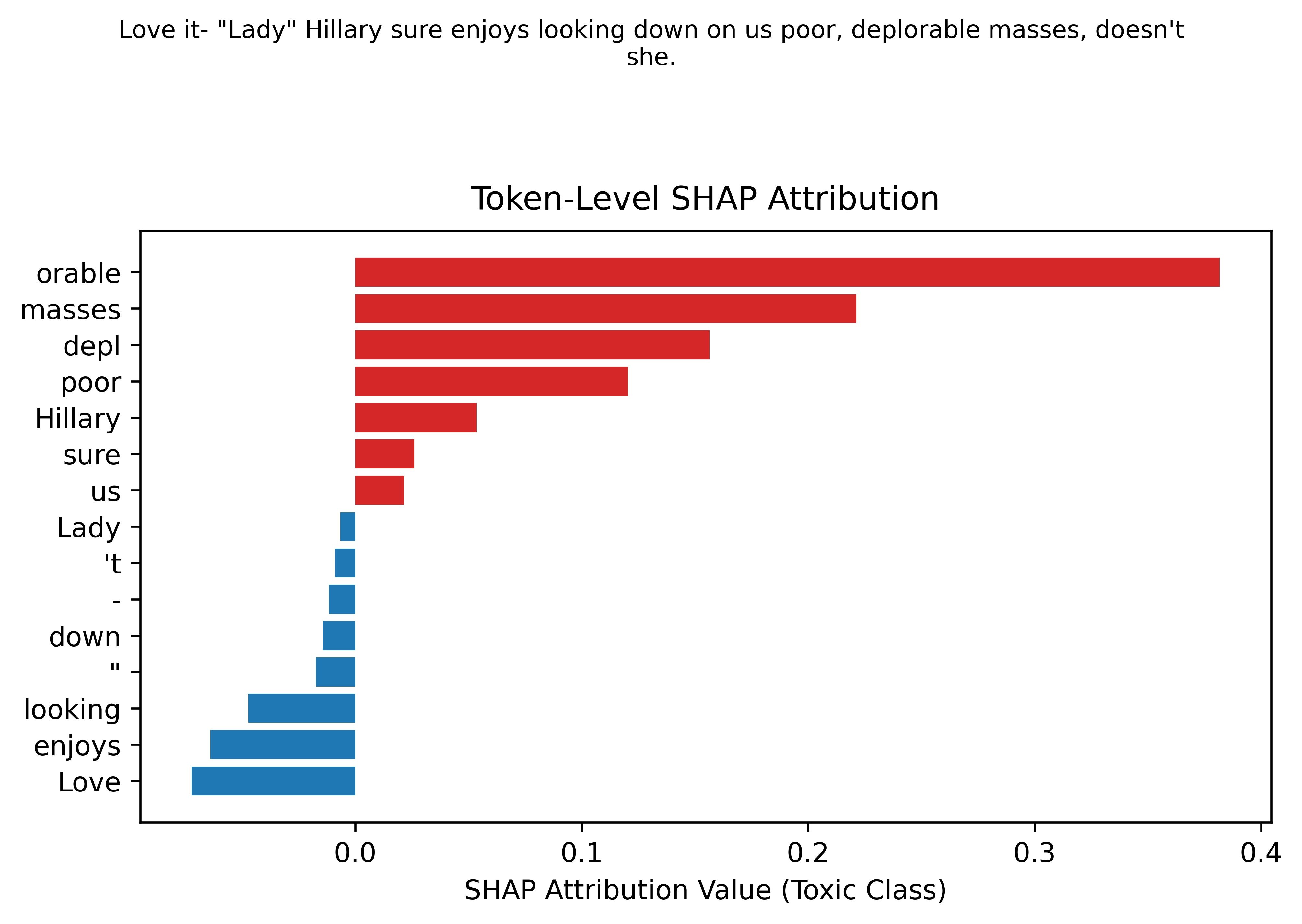}
\caption{Shapley Additive Explanations for an incorrect prediction (False Positive)}
\label{fig:shap_fp}
\end{figure}

But Shapley Additive Explanations introduce serious weaknesses on misclassified samples. For the false positive examples, the method occasionally highlights the contributions of single lexical triggers (i.e. negative words) explicitly such as swearing or political jargon, even if the triggers contained in the text do not constitute toxic behavior within the context. As presented in image~\ref{fig:shap_fp}, a SHAP approach can misclassify the example as toxic token by token, and the model behaves overconfidently and over-standardized too many examples in less aggressive situations.

\subsection{Integrated Gradients Explanations}

In contrast to Shapley Additive Explanations, Integrated Gradients produced more distributed attribution profiles which take into account cumulative contributions along the input. Rather than focusing many of the tokens of input data that have the most significant impact, it also relatively weights a broader variety of contextual and structural features. Its ability to do so is demonstrated in Table~\ref{tab:ig_tokens}, in which there is increased sensitivity to longer-range dependencies for longer comments and in the case of implicitly toxic comments. Table~\ref{tab:qualitative_compare} provides a qualitative comparison of the two attribution methods, which illustrates their differing profiles among a representative set of predictions.

\begin{table}
\caption{Top Integrated Gradients token attributions for a correctly classified non-toxic comment. The model predicts the non-toxic class with high confidence ($P(\text{non-toxic}) = 0.997$).}
\label{tab:ig_tokens}
\centering
\begin{tabular}{|l|c|}
\hline
Token & IG Attribution \\
\hline
away & 0.221 \\
war & 0.195 \\
to & 0.165 \\
firms & 0.105 \\
suspect & 0.104 \\
position & 0.100 \\
http & 0.087 \\
Energy & 0.066 \\
\hline
\end{tabular}
\end{table}

\begin{table}
\caption{Qualitative comparison of Shapley Additive Explanations and Integrated Gradients attributions on representative prediction outcomes.}
\label{tab:qualitative_compare}
\centering
\begin{tabular}{|p{3cm}|p{4cm}|p{5cm}|}
\hline
\textbf{Example Type} & \textbf{Dominant Attribution Pattern} & \textbf{Observed Behavior} \\
\hline
True Positive & Distributed across multiple tokens & Contextual cues collectively support correct toxic classification \\
False Positive & Concentrated on isolated lexical cue & Over-reliance on single trigger leads to misclassification \\
False Negative & Weak and diffuse attribution & Subtle or implicit toxicity not sufficiently captured \\
Long Comments & Broad attribution span & Captures long-range dependencies better than SHAP \\
Short Comments & Narrow attribution span & Limited context reduces explanatory clarity \\
\hline
\end{tabular}
\end{table}

Additionally, the diffuse quality of the Integrated Gradients explanations may decrease their interpretability for human beings since no one token is obviously the primary driver of the model prediction. This indicates a potential drawback of this approach in terms of the clarity of the explanations provided for the model decisions, in exchange for greater contextual sensitivity.

\section{Discussion and Conclusion}

This work conducted an explainability-driven analysis of a neural harmful content detection model. This analysis is designed to provide insights into model behavior beyond overall performance metrics and highlights how explainability methods can provide insights into important patterns and shortcomings that are not clear from standard evaluation. Despite the impressively high overall accuracy produced by the RoBERTa-based classifier, an analysis based on Shapley Additive Explanations has revealed that model predictions closely follow some explicit lexical signals. This behavior results in intuitive explanations in the correctly classified instances, but may also be responsible for misclassifications in certain subtle cases. Integrated Gradients allows to place a model attribution in a different light, distributing attributions over a larger number of tokens, while benefiting from contextual information potentially lost in sparser attributions. This is not without its drawbacks, however: increased contextual sensitivity negatively impacts interpretability because explanations become less concise and harder for human evaluators to interpret. All these findings suggest a trade-off between context coverage and clarity of explanation through post-hoc explainability methods. All in all, the findings suggest that explainable artificial intelligence can serve as a valuable mechanism for auditing and diagnosing malicious content moderation systems. By making systematic failure modes, such as lexical over-attribution and challenges in identifying indirect toxicity, visible, explanations may aid human-in-the-loop moderation and signal potential avenues for future model enhancement. This study highlights the importance of embedding explainability into the evaluation of intelligent moderation systems in order to further transparency, trustworthiness, and responsible deployment.

\subsection{Limitations}

There are some limitations for this study. Firstly, the proposed analysis is limited to one dataset and a binary classification formulation for harmful content moderation, which may not fully capture the complexity of multi-label harmful content moderation scenarios. Secondly, post-hoc model explanation approaches are sensitive to the specific tokenization, input perturbation and model non-determinism, all of which result in variability for example attribution results. Thirdly, the evaluation emphasizes qualitative interpretability rather than quantitative faithfulness metrics.  While this is justified for the diagnosed objectives taken in this study, it could be a limiting factor nonetheless. For future work, it constitutes an important step to tackle these limitations.

\begin{credits}
\subsubsection{\discintname}
The authors have no competing interests to declare that are
relevant to the content of this article.
\end{credits}
%
%
%

\begin{thebibliography}{15}

\bibitem{ref_roberta}
Liu, Y., Ott, M., Goyal, N., Du, J., Joshi, M., Chen, D., Levy, O., Lewis, M., Zettlemoyer, L., Stoyanov, V.:
RoBERTa: A robustly optimized BERT pretraining approach.
\emph{arXiv preprint arXiv:1907.11692} (2019)

\bibitem{ref_toxicmodels}
Mozafari, A., Farahbakhsh, R., Crespi, N.:
A BERT-based transfer learning approach for hate speech detection.
\emph{SN Computer Science} \textbf{1}(6) (2020)

\bibitem{ref_shap}
Lundberg, S.M., Lee, S.-I.:
A unified approach to interpreting model predictions.
In: \emph{Proceedings of the 31st International Conference on Neural Information Processing Systems (NeurIPS)}, pp.~4765--4774 (2017)

\bibitem{ref_ig}
Sundararajan, M., Taly, A., Yan, Q.:
Axiomatic attribution for deep networks.
In: \emph{Proceedings of the 34th International Conference on Machine Learning (ICML)}, pp.~3319--3328 (2017)

\bibitem{ref_civilcomments}
Borkan, D., Dixon, L., Sorensen, J., Thain, N., Vasserman, L.:
Nuanced metrics for measuring unintended bias with real data for text classification.
In: \emph{Proceedings of The Web Conference (WWW)}, pp.~491--500 (2019)

\bibitem{ref_icwsm}
Davidson, T., Warmsley, D., Macy, M., Weber, I.:
Automated hate speech detection and the problem of offensive language.
In: \emph{Proceedings of the 11th International AAAI Conference on Web and Social Media (ICWSM)}, pp.~512--515 (2017)

\bibitem{ref_naacl}
Waseem, Z., Hovy, D.:
Hateful symbols or hateful people? Predictive features for hate speech detection on Twitter.
In: \emph{Proceedings of NAACL-HLT}, pp.~88--93 (2016)

\bibitem{ref_epj}
Burnap, P., Williams, M.L.:
Cyber hate speech on Twitter: An application of machine classification and statistical modeling for policy and decision making.
\emph{EPJ Data Science} \textbf{4}(1) (2015)

\bibitem{ref_bert}
Devlin, J., Chang, M.-W., Lee, K., Toutanova, K.:
BERT: Pre-training of deep bidirectional transformers for language understanding.
In: \emph{Proceedings of NAACL-HLT}, pp.~4171--4186 (2019)

\bibitem{ref_bias}
Dixon, L., Li, J., Sorensen, J., Thain, N., Vasserman, L.:
Measuring and mitigating unintended bias in text classification.
In: \emph{Proceedings of the AAAI/ACM Conference on AI, Ethics, and Society (AIES)}, pp.~67--73 (2018)

\bibitem{ref_lime}
Ribeiro, M.T., Singh, S., Guestrin, C.:
``Why should I trust you?'': Explaining the predictions of any classifier.
In: \emph{Proceedings of the 22nd ACM SIGKDD International Conference on Knowledge Discovery and Data Mining (KDD)}, pp.~1135--1144 (2016)

\bibitem{ref_emnlp}
Arras, L., Horn, F., Montavon, G., Müller, K.-R., Samek, W.:
Explaining predictions of nonlinear classifiers in NLP.
In: \emph{Proceedings of EMNLP}, pp.~2791--2797 (2017)

\bibitem{ref_attention}
Vig, J.:
A multiscale visualization of attention in the transformer model.
In: \emph{Proceedings of the 57th Annual Meeting of the ACL (System Demonstrations)}, pp.~37--42 (2019)

\bibitem{ref_interfaces}
Amershi, S., Weld, D., Vorvoreanu, M., Fourney, A., Nushi, B., Collisson, P., Suh, J., Iqbal, S., Bennett, P., Inkpen, K., et~al.:
Guidelines for human-AI interaction.
In: \emph{Proceedings of the 2019 CHI Conference on Human Factors in Computing Systems}, pp.~1--13 (2019)

\bibitem{ref_interpret1}
Doshi-Velez, F., Kim, B.:
Towards a rigorous science of interpretable machine learning.
\emph{arXiv preprint arXiv:1702.08608} (2017)

\end{thebibliography}
%

\end{document}